\documentclass[11pt]{article}

\usepackage{acl}

\usepackage{times}
\usepackage{latexsym}

\usepackage[T1]{fontenc}
\usepackage[utf8]{inputenc}

\usepackage{microtype}

\usepackage{inconsolata}

\usepackage{graphicx}

\usepackage{amsmath}
\usepackage{booktabs}
\usepackage{multirow}
\usepackage[normalem]{ulem}
\usepackage{algorithm}
\usepackage{algpseudocode}

\title{Computational Blueprints: Generating Isomorphic Mathematics Problems with Large Language Models}

\author{
 \textbf{Jeong-Hoon Kim$^*$},
 \textbf{Jinwoo Nam$^*$}, and
 \textbf{Geunsik Jo}    
\\
 Turing Co. Ltd.
\\
 \texttt{\{eric, matthew, klip\}@teamturing.com}
}

\begin{document}
\maketitle
% 공저자 표시
\def\thefootnote{*}\footnotetext{These authors contributed equally to this work}\def\thefootnote{\arabic{footnote}}

% \begin{abstract}
% This document is a supplement to the general instructions for *ACL authors. It contains instructions for using the \LaTeX{} style files for ACL conferences.
% The document itself conforms to its own specifications, and is therefore an example of what your manuscript should look like.
% These instructions should be used both for papers submitted for review and for final versions of accepted papers.
% \end{abstract}

\begin{abstract}
Personalized mathematics education is growing rapidly, creating a strong demand for large sets of similar practice problems.
Yet existing studies on mathematics problem generation have focused on data augmentation for training neural language models rather than on direct educational deployment. 
To bridge this gap, we define a new task, Isomorphic Math Problem Generation (IMPG), designed to produce structurally consistent variants of source problems. 
Subsequently, we explored LLM-based frameworks for automatic IMPG through successive refinements, and established Computational Blueprints for Isomorphic Twins (CBIT).
With meta-level generation and template-based selective variation, CBIT achieves high mathematical correctness and structural consistency while reducing the cost of generation.
Empirical results across refinements demonstrate that CBIT is superior on generation accuracy and cost-effectiveness at scale.
Most importantly, CBIT-generated problems exhibited an error rate $17.8\%$ lower than expert-authored items, with deployment to $6,732$ learners on a commercial education platform yielding $186,870$ interactions.
\end{abstract}

% \section{Introduction}

% These instructions are for authors submitting papers to *ACL conferences using \LaTeX. They are not self-contained. All authors must follow the general instructions for *ACL proceedings,\footnote{\url{http://acl-org.github.io/ACLPUB/formatting.html}} and this document contains additional instructions for the \LaTeX{} style files.

% The templates include the \LaTeX{} source of this document (\texttt{acl\_latex.tex}),
% the \LaTeX{} style file used to format it (\texttt{acl.sty}),
% an ACL bibliography style (\texttt{acl\_natbib.bst}),
% an example bibliography (\texttt{custom.bib}),
% and the bibliography for the ACL Anthology (\texttt{anthology.bib}).

\section{Introduction}
In mathematics education, repetitive practice with similar problems is essential for conceptual understanding, as demonstrated by decades of research.
According to \citet{gickSchemaInductionAnalogical1983}, students learn by comparing and contrasting multiple similar examples.
Through this process, they abstract the core structure of problems, which is the key to transfer knowledge to novel situations.
Similarly, \citet{rittle2001} highlights the link between procedural skill and conceptual insight.
Procedural skill, which is strengthened through repeated practice, develops into procedural fluency that reduces the cognitive load required for computation and allows students to focus on underlying concepts.

Despite its educational importance, providing high-quality repetitive learning experiences remains a formidable challenge in industrial settings.
Generating related practice problems has been regarded as the work of expert mathematics educators, since every new item must remain pedagogically valid and tightly aligned with the targeted concepts \citep{burgos2025model}, demanding substantial time and cost.
In personalized online learning environments, the challenge becomes even greater. 
Because the required practice problems differ depending on the student and the topic, it is impractical to prepare them all in advance.

In a different line of work, automatic generation of mathematics problems has been an active research area in Neural Language Processing with a variety of approaches. \citet{zhou-huang-2019-towards}, \citet{liu-etal-2021-mathematical}, \citet{wang-etal-2021-math}, \citet{wu2022}, and \citet{qin2023mathematical} focused on effectively encoding equation and context information into neural language models for mathematics problem generation, but the range of producible problems was severely limited by the capacity of those small models.
The advent of the Transformer architecture \citep{vaswani2017attention} and the scaling-law \citep{kaplan2020scaling} motivation for Large Language Models (LLMs) broadened the field.
For example, \citet{drori2022} employed the Codex \citep{chen2021evaluating} to synthesize new questions by leveraging other problems within the same curricular unit, but provided no guarantee of the mathematical correctness of the generated items.
\citet{li2024neuro} projected original problems into a symbolic domain (SMT-LIB; \citealp{SMT-LIB}), where projection and reconstruction were performed by GPT-4\citep{achiam2023gpt} and symbolic mutation were applied to generate similar problems, ensuring mathematical correctness using SMT-Lib solver.
Similarly, the AIC method \citep{li2024synthesizing} used Llama3-70B-Instruct \citep{grattafiori2024llama} to project and reconstruct problems in an abstract domain, while also prompting the model to generate verification programs that check the abstraction.
Both studies provided significant insight to this work in that they transformed the task domain so that mathematical correctness can be verified deterministically. Specifically, AIC, which supports a broader range of problem types, was adopted as the baseline for comparison; further details are given in Section~\ref{sec:proposed_method}.

Despite these advances, no prior work has explicitly aimed at educational use.
The above methods evaluate performance by semantic similarity measured with BLEU \citep{papineni-etal-2002-bleu}, ROUGE \citep{lin-2004-rouge}, and METEOR \citep{banerjee-lavie-2005-meteor} between generated problems and their input equations or context, and by downstream benchmark scores on mathematics problem solving such as GSM8k \citep{cobbe2021training} and MATH \citep{chen2021evaluating}. 
This evaluation scheme indicates that existing studies concentrated on data-augmentation technique for training neural language models, and in fact present their work as mathematical data synthesis.
Consequently, deploying such problems for educational use incurs substantial hidden costs for human expert verification, ultimately nullifying the benefits of automation.

\begin{figure*}[h!]
  \centering
  \includegraphics[width=0.65\textwidth]{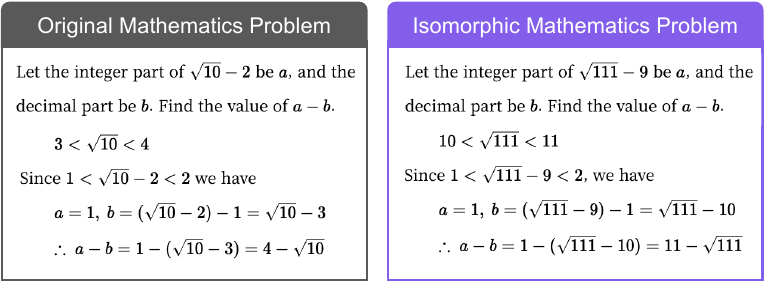}
  \caption {Example of an isomorphic math problem generated from an original one, showing changed numerical values while preserving the underlying structure.}
  \label{fig:iso}
\end{figure*}

To bridge the gap between academic research and educational application, we first propose the task of Isomorphic Math Problem Generation (IMPG).
As shown in Figure \ref{fig:iso}, rather than inventing entirely new problems, IMPG aims to generate structurally consistent variants of source problems that vary only in numerical values and thereby inherit the pedagogical validity of the source problems’ linguistic context. 
Therefore, if the source problems have been pedagogically validated, the isomorphic problems successfully generated through IMPG can in turn be considered valid.
To facilitate research on this task, we construct and publicly release\footnote{\url{https://github.com/eric-jhoon/IMPG}} a comprehensive benchmark dataset covering multiple grade levels and difficulty tiers, accompanied by an automatic verification toolkit that symbolically checks whether generated problems satisfy the required numerical relationships, thereby enabling reliable and standardized evaluation of new methods.

To address IMPG, we propose Computational Blueprints for Isomorphic Twins (CBIT), a framework designed to ensure mathematical correctness, structural consistency, and scalability.
For mathematical correctness, CBIT encodes the quantitative relationships of source problems as symbolic constraints and generates new items only when those constraints are satisfied.
For structural consistency, it isolates invariant linguistic and symbolic templates from variable numerical elements, filling only designated slots so that each output preserves the source form.
For scalability, CBIT adopts a meta-level approach in which the LLM writes a reusable problem-generating program, enabling unbounded verified outputs with minimal LLM calls.

We summarize our contributions as follows.
\begin{itemize}
    \item We propose the task of Isomorphic Math Problem Generation (IMPG), aimed at creating problems ready for real educational deployment.
    \item We build and release a benchmark dataset covering diverse grade levels and difficulty ranges, accompanied by an automatic verification toolkit that enables reliable evaluation and facilitates future research on IMPG.
    \item We establish CBIT framework that leverages LLMs as blueprint authors to achieve high generation accuracy and substantially lower production cost, while aligning LLM capabilities with the requirements of IMPG.
    \item We validate CBIT’s industrial applicability through a large-scale real-world deployment, where it achieved an error rate 17.8\% lower than expert-authored items over 186,870 learning interactions involving 6,732 learners, demonstrating its readiness for practical use in education at scale.
\end{itemize}

\section{Isomorphic Mathematics Problem Generation}
\subsection{Task Definition}
We define a mathematics problem as a tuple $P=(Q,S,A)$, where
$Q$ is the question text, $S$ is the solution text, and $A$ is the final answer entailed by $S$ for $Q$.
Note that each of Q and S may consist of natural language, mathematical expressions, or a mixture of both, and must contain at least one mathematical expression. 
The goal of Isomorphic Math problem Generation is to generate a set of $K$ problems $\mathcal{T}=\{P_1,P_2, \dots,P_K\}$ from a source problem $P_s$ where each $P_i$ in $\mathcal{T}$ satisfies the following conditions.

\textbf{Structural Equivalence}: These conditions ensure that any variation is restricted to the content of the original mathematical expressions, preserving the structural essence of the source problem.
\begin{align*}
\begin{aligned}
&\hspace*{-1.5em}\forall i \in \{1,\dots,K\}: \\
&E(Q_s \oplus S_s)\neq E(Q_i \oplus S_i),\\
&|E(Q_s \oplus S_s)| = |E(Q_i \oplus S_i)|,\\
&N(Q_s \oplus S_s) = N(Q_i \oplus S_i),
\end{aligned}
\end{align*}
where $\oplus$ denotes string concatenation, $E(\cdot)$ and $N(\cdot)$ returns the tuple of all mathematical expressions and natural-language text contained, respectively. $|\cdot|$ represents the number of elements in a given tuple.

\textbf{Conceptual Correctness}: 
These conditions ensure that every generated problem preserves the quantitative and logical relationships
among the mathematical expressions of the source problem, thereby maintaining the original mathematical concepts.
\begin{align*}
\begin{aligned}
&\hspace*{-1.5em}\forall i \in \{1,\dots,K\}: \\
&\mathcal{R}\big(E(Q_s \oplus S_s)\big)
   = \mathcal{R}\big(E(Q_i \oplus S_i)\big),
\end{aligned}
\end{align*}
where $\mathcal{R}(\cdot)$ denotes the implicit ideal set of quantitative relationships
among numerical values, encompassing dependencies both within and across expressions.
Only when all the above conditions are satisfied is the Isomorphic Math Problem Generation (IMPG) task considered successfully accomplished; 

\subsection{Benchmark}
\begin{table}
\centering
% \resizebox{\columnwidth}{!}{%
\resizebox{0.7\linewidth}{!}{
\begin{tabular}{l r r r}
\toprule
Difficulty & Total Rel & Avg. & Max \\
\midrule
Easy   &   605  &  9.17 & 28 \\
Medium &   832  & 12.42 & 29 \\
High   &   973  & 15.69 & 35 \\
\midrule
Total  & 2,410  & 12.36 & 35 \\
\bottomrule
\end{tabular}}%
% }
% \caption{Symbolic relationships in the dataset by difficulty level.}
\caption{Symbolic relationships counts by difficulty level. 
Total Rel indicates the total number of relationships identified, while Avg. and Max denote the average and maximum number of relationships per problem, respectively.}
\vspace{-1em}
\label{tab:difficulty-relations}
\end{table}
A benchmark dataset was constructed to support the evaluation presented in this study, covering mathematics problems from grade 4 through grade 12 (up to college entrance exam level). This dataset comprises $195$ original problems sampled across $67$ distinct curriculum units, with each problem categorized into low, medium, or high difficulty levels. 
Specifically, The dataset was sampled from a pool of expert-authored and field-tested items, ensuring both pedagogical validity and practical relevance. We then applied an additional restriction to include only text-based items represented in LaTeX, while maintaining balanced coverage across curriculum units and difficulty levels.

Note that under IMPG, the dataset contains far richer information than the raw count suggests.
As the problem generation is reframed into revealing symbolic relations between variables, our dataset encodes 2,410 relation signals rather than merely 195 isolated items.
As shown in Table~\ref{tab:difficulty-relations}, the average number of identifiable symbolic relationships per problem increases steadily from easy to hard, and the dataset as a whole contains 2,410 such relationships.
A detailed analysis of the dataset distribution and characteristics is provided in Appendix \ref{sec:appendix_dataset}. 

Unlike conventional datasets providing a fixed set of gold answers, we instead develop and release a problem-specific verification toolkit. This toolkit, implemented as a lightweight algorithm, enables the automatic validation of any number of generated isomorphic problems. The example pseudo-code of the toolkit is provided in Appendix Algorithm~\ref{alg:verification}. Since evaluation in this work is conducted not by comparing against pre-defined gold answers but by applying automatic verification algorithms, the practical evaluation scale is substantially amplified. For instance, if each of the $195$ problems is used to generate $10$ isomorphic problems, the evaluation effectively covers $1,950$ problem instances. This design enables large-scale, rigorous assessment without requiring manual annotation.

\subsection{Metric}
The major evaluation metric is the Mean Per-Problem Success Rate(MPSR). Given a total of $N$ original problems, for each original problem indicator $i$, the generation of a fixed number $K$ of isomorphic problems would be requested (e.g., $K=30$). Let $s_i$ denote the number of successfully generated problems that pass verification for problem $i$, where $0\leq s_i\leq K$. The per-problem success rate is then defined as:
\begin{equation}
  \label{eq:per-problem success rate}
  \mathrm{SuccessRate}_i = \dfrac{s_i}{K},
\end{equation}
that is, the ratio of successful generations relative to the requested number of generations. The overall MPSR is computed by taking the mean of the per-problem success rates across all $N$ problems:
\begin{equation}
  \label{eq:mpsr}
  \mathrm{MPSR} = \dfrac{1}{N}\sum_{i=1}^{N}\dfrac{s_i}{K}
\end{equation}

In addition to success rates, we also evaluate the economic efficiency of problem generation using token consumption. Let $T_{pr}$ and $T_{cp}$ denote the total number of prompt tokens and completion tokens used, respectively. We define Tokens Per Success(TPS) as the total token consumption divied by the number of successful generations. To express this as a real-world cost in USD, we incorporate the per-token costs of prompt tokens $c_{pr}$ and completion tokens $c_{cp}$, which may differ depending on the LLM service provider. The resulting metric, Cost Per Success(CPS), is computed as:
\begin{equation}
  \label{eq:tps}
  \mathrm{CPS} = \dfrac{c_{pr} T_{pr} +c_{cp}T_{cp}}{\sum_{i=1}^{N}s_i}
\end{equation}
This cost-centric metric captures the practical efficiency of each generation method and reflects its commercial viability for real-world deployment.

\section{Proposed Method}
\label{sec:proposed_method}
\subsection{Iterative vs. Batch}
In automated math problem generation, two prompting strategies for Large Language Models (LLMs) can be leveraged; the Iterative approach and the Batch approach. The Iterative approach generates one problem per prompt and repeats the process multiple times to produce multiple problems. In contrast, the Batch approach requests LLM to generate multiple problems within a single response. 

To explore the most suitable prompting strategy for IMPG, we compared two approaches, where detailed experimental results are provided in Appendix~\ref{sec:appendix_ib}. This comparison revealed three key findings:
(1) as the request size \(K\) increased, the LLM’s individual responses in the Batch setting becomes longer and LLM tends to omit the part of the requested output, 
(2) the Iterative method, requiring a fresh prompt for every problem while the prompt itself is typically much longer than the generated answer tokens, incurred a disproportionately high token cost, and 
(3) all Batch-generated problems originating from the same source either jointly passed or jointly failed the benchmark verification, suggesting that the Batch method forms an internal generation engine before sampling individual problems. 

Taken together, these factors underscore that the Batch approach offers better scalability and internal consistency for IMPG, motivating our decision to adopt it as the baseline for all subsequent experiments.

\begin{figure}[t]
  \includegraphics[width=\columnwidth]{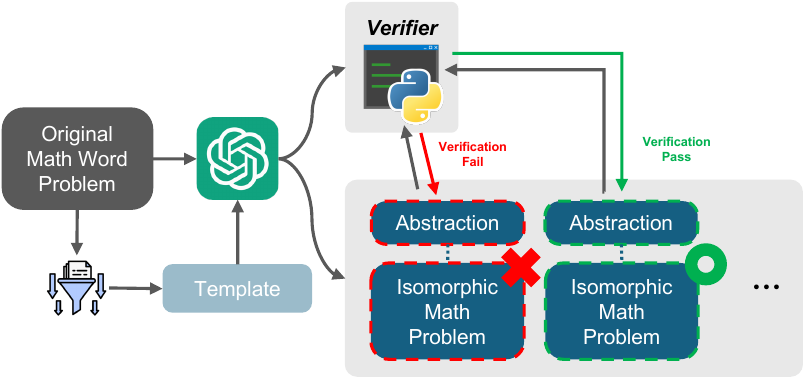}
  \caption{Structure of AIC-Batch. The LLM generates multiple Abstraction-Problem sets and a Verifier to ensure mathematical correctness, while a heuristically extracted template reinforces structural consistency.}
  \label{fig:aicb}
\end{figure}

\subsection{AIC-Batch}
To establish this baseline, we reimplemented the AIC framework, which is most closely aligned with our task, using the Batch prompting strategy.

Specifically, following the original AIC framework, the LLM is prompted to generate not only the math problem itself but also an accompanying Abstraction and Verifier code to ensure mathematical correctness. The Abstraction serves as a symbolic summary that captures key numerical elements and the answer, while the Verifier is a Python program that checks the consistency of relationships between these values. 

In addition, to further strengthen structural consistency, we heuristically extracted a natural text template from each source problem and provided it to the LLM as part of the prompt, so that the generated problems would adhere more tightly to the intended linguistic and structural patterns.

This design strengthens the core requirements of IMPG, namely mathematical correctness and structural consistency, thereby providing a robust baseline built on the strengths of the original AIC method.

\subsection{Blueprint for Isomorphic Twin}
\begin{figure}[t]
\centering
\resizebox{0.9\linewidth}{!}{
  \includegraphics[width=\linewidth]{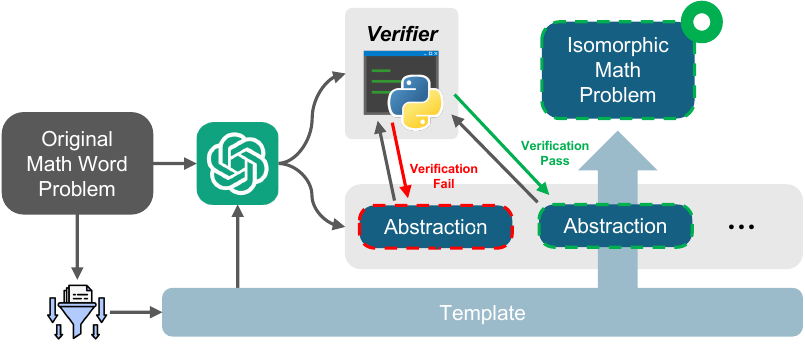}
}
  \caption{Blueprint for Isomorphic Twin. The LLM outputs only an Abstraction while a predefined template deterministically reconstructs the full text.}
  \vspace{-0.7em}
  \label{fig:bit}
\end{figure}

Motivated by our earlier finding that the LLM tends to avoid long responses, we explored whether further improvement could be achieved by having the model generate only the Abstraction.
If the LLM were limited to this concise symbolic representation, the output length would shrink substantially, which we expected would mitigate the long-response avoidance and at the same time free the model to concentrate on mathematical reasoning.
However, such a design requires a way to recover the full text from the Abstraction.
To meet this requirement, we developed problem templates capable of deterministically combining with the generated Abstraction to reconstruct complete question and solution texts.
This blueprint–template design defines our Blueprint for Isomorphic Twin (BIT) framework.

By explicitly decomposing problem generation into Abstraction generation and text realization, BIT assigns only the semantically demanding symbolic reasoning to the LLM while the natural-language layer is completed heuristically from the template.
This not only reduces the cognitive and computational burden of free-form text production but also yields two additional advantages.
First, whereas AIC-Batch merely provided templates as optional references, BIT enforces template-guided generation, allowing much stricter structural consistency than before.
Second, because the full text is now constructed deterministically from the verified Abstraction, mathematical correctness established at the Abstraction level automatically extends to the final question and solution text.
In contrast, the earlier AIC-Batch setting left the Abstraction and full text only loosely coupled, so correctness of the Abstraction did not strictly guarantee correctness of the realized problem.
% Figure \ref{fig:aicb} and Figure \ref{fig:bit} illustrate how the relationships among the Verifier, Abstraction, and Isomorphic Math Problem differ between the two designs.
A comparison of Figure~\ref{fig:aicb} and Figure~\ref{fig:bit} makes clear the differing relationships among the Verifier, Abstraction, and Isomorphic Math Problem in the two designs.

Furthermore, BIT can be viewed as externalizing a latent two-stage plan that the LLM was already performing implicitly.
In the Batch setting we had observed that all generated variants of a given source problem either jointly passed or jointly failed the benchmark verification, suggesting that the model was already forming an internal numerical blueprint before realizing the surface text.
By making this hidden planning explicit and restricting the model to the essential subtask of formula generation, BIT enables more stable and scalable isomorphic problem generation while directing the model’s computational effort toward the core symbolic reasoning required for the task.

\subsection{Computational Blueprint for Isomorphic Twin}
\begin{figure}[t]
\centering
\resizebox{0.9\linewidth}{!}{
  \includegraphics[width=\linewidth]{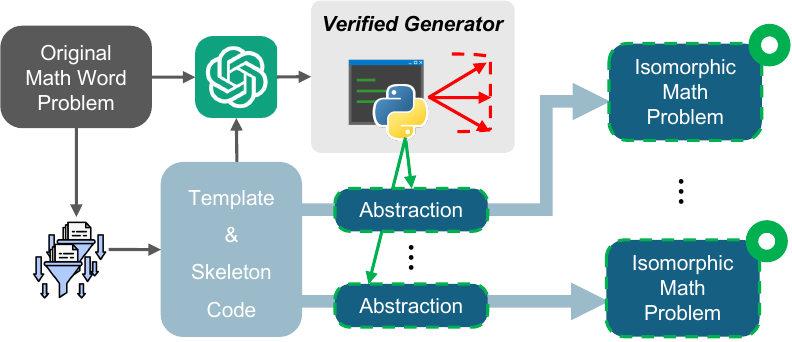}
}
  \caption{Computational Blueprint for Isomorphic Twin. The LLM outputs a verified generator program that produces Abstractions from random seeds and full isomorphic problems is reconstructed by predefined template.}
  \vspace{-0.7em}
  \label {fig:cbit}
\end{figure}

Building on BIT, we sought to push task decomposition one step further.
While BIT externalizes the latent separation between abstraction generation and text realization, we asked whether the Abstraction generation itself could also be externalized.
Our earlier observations suggested that, in the Batch setting, the LLM already forms an implicit generation engine before producing text.
This led us to investigate whether the model could be prompted to explicitly create such an engine as an executable program, rather than directly emitting problems or even their Abstractions.

In the resulting framework, called Computational Blueprint for Isomorphic Twin (CBIT), the LLM is no longer instructed to output problem text or Abstractions.
Instead, it is prompted to write a generator program that takes a random seed as input and, for each seed, produces an isomorphic math problem together with its solution and answer.
Because program synthesis is inherently more free-form than producing a concise Abstraction, we provide a skeleton code specifying essential interfaces and control flow to guide the LLM toward consistent and verifiable outputs. Detailed description of code-guided schema is described in Appendix \ref{sec:appendix_cg}. This reframes problem generation as a meta-generation task where the LLM’s effort is directed toward constructing the generative mechanism itself.

The design of CBIT yields several important advantages.
First, verification becomes intrinsic to generation.
Because the generator program is required to construct Abstractions that already satisfy all numerical relationships identified in the source problem, mathematical correctness is guaranteed by deterministic execution, eliminating the need for a separate Verifier and avoiding repeated LLM calls whenever verification fails.
Consequently, the number of generated problems becomes independent of LLM invocation cost; once the generator is authored, it can produce any number of correct variants with negligible additional expense. Second, by shifting the production of Abstractions from free-form text to code-based computation, CBIT strengthens mathematical correctness.
Prior work such as Program-of-Thought \cite{chen2023program} has shown that when LLMs are guided to reason via external tools or code, their mathematical reliability improves.
CBIT exploits this effect by letting the LLM use its reasoning ability to design the generator while delegating the exact calculation of numerical values to a deterministic programming language, ensuring computational integrity. Third, CBIT offers an exceptional level of expert controllability and interpretability.
Because the LLM’s output is a generator program, the logic by which new problems are created is fully transparent and can be readily inspected by human experts.
If any generated problems are deemed mathematically unsuitable, minor edits to the program suffice to regenerate the entire family of isomorphic problems without further LLM involvement.
Moreover, storing the generator itself provides a long-term advantage; the program can be preserved and re-executed at any time to produce additional verified problems on demand, ensuring reproducibility and efficient maintenance of large problem banks.

In summary, CBIT transforms the role of the LLM from producing individual problems to authoring a computational generator, thereby combining the strengths of symbolic reasoning and deterministic execution.  
It achieves built-in mathematical correctness, decouples generation scale from token cost, and provides strong expert controllability and long-term reproducibility.

\section{Experimental Results}

\begin{table}
  \centering
    \resizebox{0.75\linewidth}{!}{
    \begin{tabular}{lrccc}
    \toprule
    Metric & K & AIC-B & BIT & CBIT \\
    \midrule
    \multirow{3}{*}{MPSR↑}
      & 10  & 0.39 & \uline{0.85} & \bfseries 0.93 \\
      & 30  & 0.17 & \uline{0.79} & \bfseries 0.89 \\
      & 100 & 0.01 & \uline{0.32} & \bfseries 0.83 \\
    \midrule
    \multirow{3}{*}{CPS↓}
      & 10  & \$17.90 & \uline{\$7.27} & \bfseries \$6.28 \\
      & 30  & \$12.53 & \uline{\$3.41} & \bfseries \$2.17 \\
      & 100 & \$30.03 & \uline{\$1.58} & \bfseries \$0.71 \\
    \bottomrule
    \end{tabular}
    }
    \caption{Comparison of MPSR, and CPS across different request sizes. CPS is measured in USD per 1,000 problems. Best in \textbf{bold}, second-best \uline{underlined}.}
  \label{tab:MPSR}
\end{table}

\textbf{Generation Accuracy and Cost Effectiveness:}
We compare the three proposed isomorphic problem generation frameworks using Mean Per-Problem Success Rate (MPSR), and Cost Per Success (CPS) as evaluation metrics. As shown in Table~\ref{tab:MPSR}, across all metrics and request sizes, the three methods improve in clear stages demonstrating that each design choice contributed in the intended direction.

From AIC-Batch to BIT, the introduction of the blueprint mechanism markedly strengthened structural consistency and concentrated the LLM's effort on mathematical reasoning. This is reflected in the large jump in MPSR; even at $K=30$, BIT achives  $0.79$ compared to AIC-Batch's $0.17$.
Notably, CBIT's MPSR remains close to BIT's up to $K=30$, indicating that the blueprint is the dominant driver of generation accuracy, while CBIT provides additional but more modest gains in this dimension.

In contrast, cost efficiency improves most dramatically when moving from BIT to CBIT.
At $K=100$, CBIT achieves a CPS of $\$0.71$, improving on BIT's $\$1.58$ and representing a $42\times$ lower cost than AIC-Batch's $\$30.03$.
This striking decrease stems from CBIT’s computational approach, where the LLM writes a generator program that can produce an unlimited number of verified problems without further model calls, thereby decoupling generation scale from token cost.

\begin{figure}[t]
\begin{center}
  \includegraphics[width=7cm]{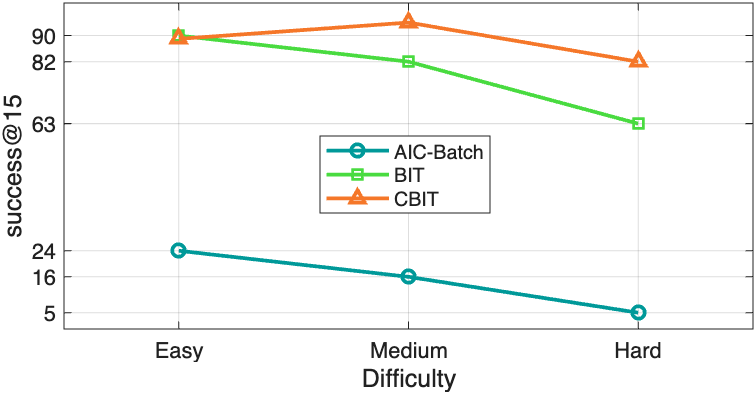}
  \caption {Success@15 across difficulty levels}
  \label{fig:difficulty}
\end{center}
\vspace{-2em}
\end{figure}

\textbf{Robustness across Difficulty Levels}:
According to Table~\ref{tab:difficulty-relations}, increasing problem difficulty corresponds to a greater number of numerical relationships, which in turn introduces two distinct obstacles; the generation process may fail to identify all necessary relationships, or, even when these relationships are successfully identified, it may fail to generalize them and instead impose constraints that are too tight, leaving too few feasible variants to meet the requested number.

To assess robustness to problem difficulty levels, we adopt the success@15 metric under the setting of generating $K=30$ isomorphic problems per source problem, with difficulty categorized as easy, medium, or hard by experts. 
This metric measures the probability that at least 15 of the requested 30 problems are successfully generated, so that cases producing only a small number of valid problems far below the request can be regarded as unsuable.

As shown in Figure~\ref{fig:difficulty}, AIC-Batch exhibits poor performance across all levels, with success@15 values of only $24\%$, $16\%$, and $5\%$ respectively, confirming its vulnerability to increased difficulty. In contrast, BIT significantly improves stability, achieving $90\%$, $82\%$, and $63\%$ across the same levels.

CBIT delivers the best and most consistent results, with success@15 reaching $89\%$, $94\%$, and $82\%$ even for the highest difficulty. CBIT demonstrates minimal sensitivity to difficulty changes, indicating strong reliability without reliance on external verifiers.

\textbf{Industrial Readiness}
We conduct a 5 months, large-scale deployment of CBIT-generated problems in a live educational service. The key metric for industrial readiness is content error rate, defined as the percentage of served problems that were reported by users and required correction. We measured it against a baseline of isomorphic problems created by human experts.

In this deployment, 32,131 problems generated by CBIT were served to 6,732 students, resulting in a total of 186,870 problem-solving attempts. For the control group, 4,404 human-authored problems were served to 3,976 students, leading to 103,231 attempts. The total pool of problems available for the experiment consisted of 195,941 from CBIT and 7,145 from human experts.

The CBIT-generated content demonstrated an error rate of 0.1867\%. In contrast, the human-authored content had a slightly higher error rate of 0.2271\%. This outcome demonstrates that CBIT can produce educational content at scale with a level of reliability on par with, or even superior to, that of human experts.

\section{Conclusion}
We introduced Isomorphic Math Problem Generation (IMPG) as a new task aimed at creating reliable practice problems suitable for real educational deployment, and released a benchmark dataset of 195 source problems with 2,410 symbolic relationships together with an automatic verification toolkit for reproducible evaluation.

Through a series of progressive frameworks, we systematically aligned large language models with the structural and mathematical requirements of IMPG, culminating in Computational Blueprint for Isomorphic Twin (CBIT), which guarantees built-in correctness, decouples generation scale from token cost, and provides strong expert controllability and long-term reproducibility.

Extensive experiments and a real-world deployment to 6,732 learners with 186,870 interactions demonstrated that CBIT achieves the highest success rates and lowest cost, reducing error rates by 17.8\% compared to expert-authored problems and establishing a robust foundation for scalable, verifiable educational content generation.

\section*{Limitations}
While our framework demonstrates strong scalability and reliability, certain assumptions underline its design.  
The construction of templates and skeleton code has thus far relied on heuristics, which must be specified with care for each application domain; for example, in our setting mathematical expressions were assumed to appear in LaTeX notation delimited by dollar signs, whereas domains with different conventions may require more elaborate parsing strategies.  
Moreover, the approach builds variants by altering numerical parameters, which is most effective for problems where quantitative changes directly influence the solution path or answer (such as word problems with rates, ratios, or probabilities, geometry with numeric measures, and equation-based items with perturbed coefficients).  
Tasks whose correctness depends solely on logical structure and is largely insensitive to numeric variation, such as truth-teller or liar puzzles, fall outside the current scope.  
Exploring ways to relax these assumptions, extend CBIT to broader classes of problems, and automate template construction more systematically represents a valuable direction for further study.

\clearpage
\appendix
\onecolumn
\section{Details on Benchmark}
\label{sec:appendix_dataset}

\begin{table}[h!]
\centering
\begin{tabular}{@{}p{0.3\textwidth} p{0.3\textwidth} p{0.3\textwidth}@{}}
\toprule
\textbf{Grades 4–6} & 
\textbf{Grades 7–9} & 
\textbf{Grades 10-12} \\ 
\midrule
Angles; Pattern Finding; Addition/Subtraction of Decimals; Polygons; \newline
Multiplication and Division; Addition/Subtraction of Fractions; Quadrilaterals; \newline
Mixed Operations with Natural Numbers; Multiples and Factors; Patterns and Correspondence; \newline
Simplification and Common Denominators; Perimeters and Areas of Polygons; \newline
Division/Multiplication of Fractions and Decimals; Ratio and Proportion; Various Graphs; \newline
Surface Area and Volume of Rectangular Prisms; Proportions and Distribution; Averages and Probability; Measurements
&
Prime Factorization; Integers and Rational Numbers; Rational Numbers and Repeating Decimals; \newline
Algebraic Expressions; Coordinate Plane and Graphs; Linear Inequalities and Systems; \newline
Graphs of Linear Functions; Quadratic Equations; Quadratic Functions; \newline
Basics of Geometry; Plane Figures; Similarity and Pythagorean Theorem; \newline
Probability; Trigonometric Ratios; Statistics; Data Representation and Interpretation
&
Polynomials; Equations and Inequalities; Equations of Figures; Sets and Propositions; \newline
Functions and Graphs; Counting Principles; Sequences; Exponential and Logarithmic Functions; \newline
Trigonometric Functions; Differentiation of Polynomial Functions; Limits of Sequences and Functions; \newline
Integration of Polynomial Functions; Probability; Statistics; Conic Sections; \newline
Integration; Permutations and Combinations; Plane Vectors; Solid Geometry and Coordinates \\
\bottomrule
\end{tabular}
\caption{Curriculum coverage of the benchmark dataset. 
The topics were selected based on the official Korean national curriculum, ensuring alignment with the progression of regular school education. 
When the same topic name appears across different grade ranges, the content reflects a deeper or more advanced treatment appropriate for that grade level.}
\label{tab:curriculum-3col}
\end{table}

\begin{figure}[h!]
  \includegraphics[width=\textwidth]{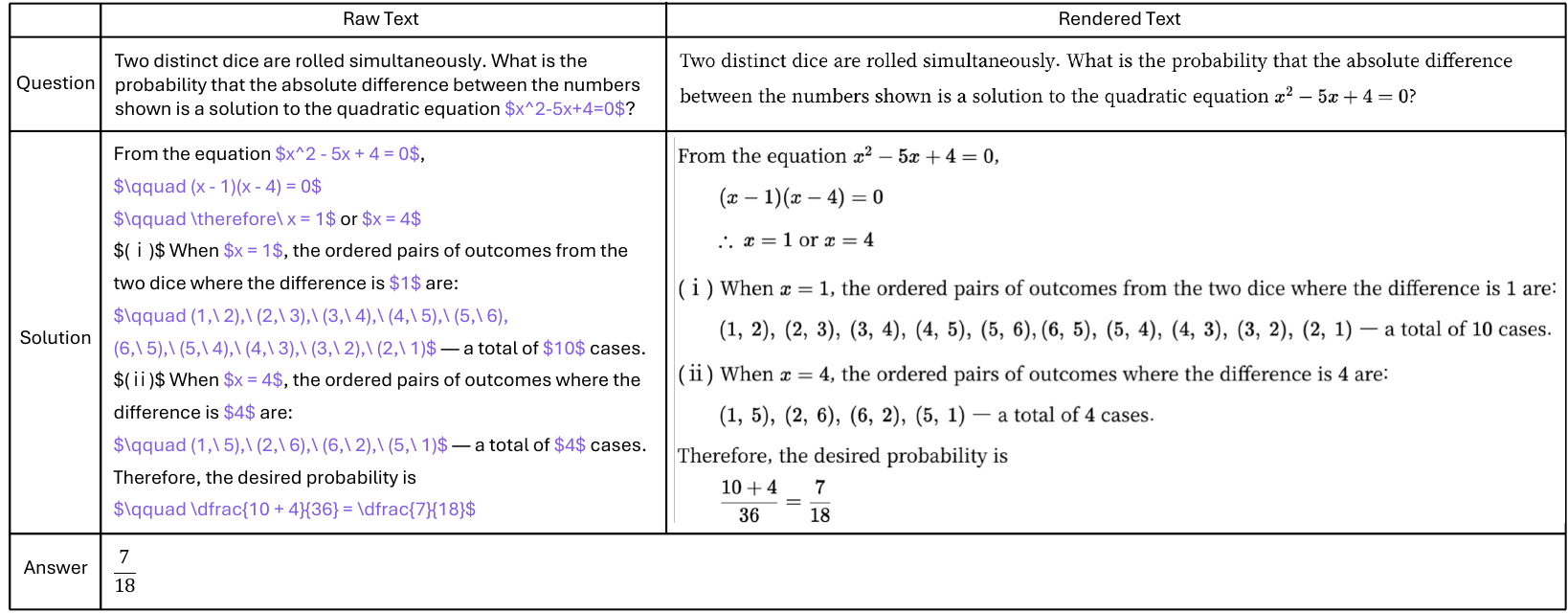}
  % \caption{Problem Sample}
  \caption{Example of a problem instance in the dataset (Problem \#20570, Grade 7 quadratic equation). 
The Raw Text column shows the original LaTeX input as stored in the dataset, while the Rendered Text column shows the corresponding human-readable rendering. 
Both the question and solution are included, together with the final answer. 
This format ensures that problems can be consistently presented to learners while also enabling symbolic parsing for verification in IMPG.}
  \label{fig:problem_sample}
\end{figure}

\clearpage

\begin{table}[h!]
\centering
\begin{tabular}{r l r r r}
\toprule
Rank & Relationship Type        & Count & Ratio   & Cumulative \\
\midrule
1  & Equality                 & 1513 & 62.78\% & 62.78\% \\
2  & Multiplicative           &  308 & 12.78\% & 75.56\% \\
3  & Additive                 &  264 & 10.95\% & 86.51\% \\
4  & Additive, Multiplicative &   81 &  3.36\% & 89.87\% \\
5  & Count                    &   52 &  2.16\% & 92.03\% \\
\vdots & \vdots               & \vdots & \vdots & \vdots \\
44 & Power, Modulo           &    1 &  0.04\% & 100.00\% \\
\bottomrule
\end{tabular}
% \caption{Distribution of relationship types in the dataset.}
\caption{Distribution of symbolic relationship types identified in the dataset. 
Each problem is decomposed into its underlying quantitative relationships, and their frequencies are aggregated across all 195 source problems. }
\label{tab:relationship-types}
\end{table}

\begin{algorithm}[h!]
\caption{Verification Procedure for Problem \#20570}
\begin{algorithmic}[1]
\Require Isomorphic Problem of \#20570
\Ensure Verification Result

\State Extract all mathematical expressions from the problem and solution texts
\State Identify the quadratic equation of the form $x^2 - ax + b = 0$
\State Identify the two roots from the from $(x-r_1)(x-r_2)=0$
\State Check that the roots satisfy:
  \Statex \hspace{2em} $r_1 + r_2 = a$, and $r_1 \cdot r_2 = b$

\For{each root $r$}
  \State Identify the list of pairs $L_r$ as written in the solution
  \State Check that each $(a, b)$ in $L_r$ satisfies:
    \Statex \hspace{2em} (i) $1 \leq a, b \leq 6$,
    \Statex \hspace{2em} (ii) $|a - b| = r$,
    \Statex \hspace{2em} (iii) all pairs are unique and distinct
  \State Check that the reported count of pairs equals $|L_r|$
\EndFor

\State Check that the total number of favorable outcomes $|L_{r_1}| + |L_{r_2}|$ is divided by $36$
\State Check the computed value is correctly simplified to a fraction or decimal
\State \Return \texttt{True} if all checks are satisfied; otherwise, \texttt{false}
\end{algorithmic}
\label{alg:verification}
\end{algorithm}

\noindent
Example verification procedure for an isomorphic problem. 
The algorithm illustrates how symbolic checking is performed: extracting expressions, validating algebraic relationships, verifying solution pairs, and ensuring correctness of the reported probability. 
This procedure exemplifies the automatic verification toolkit used to evaluate generated problems in IMPG.

\clearpage
\section{Detailed Comparison on Iterative and Batch prompting}
\label{sec:appendix_ib}

\begin{figure}[h!]
\begin{center}
  \includegraphics[width=5cm]{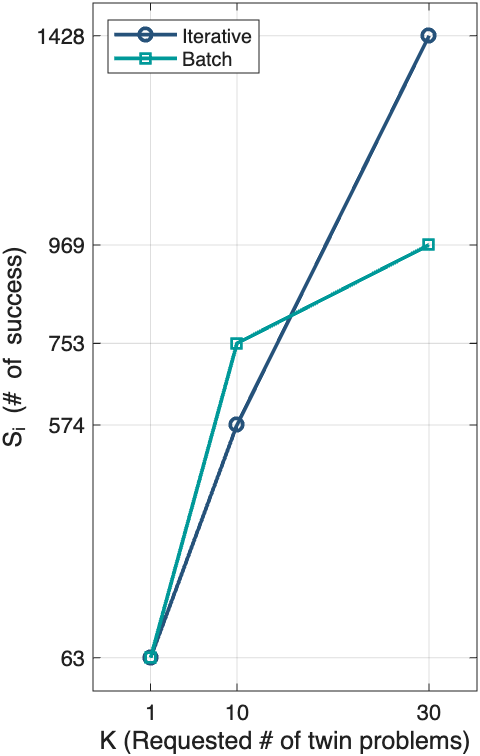}
  \includegraphics[width=5cm]{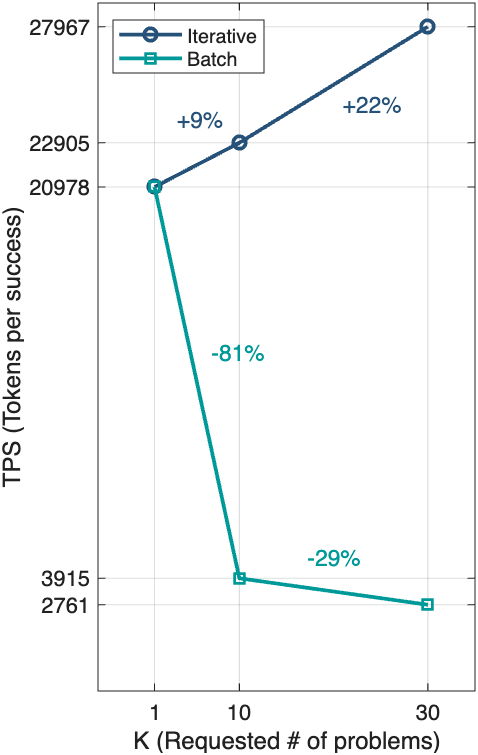}
  \caption {Comparison between Iterative and Batch prompt for IMPG with benchmark Dataset.}
  \label{fig:iterative_vs_batch}
\end{center}
\end{figure}

% As shown in the left plot, both approaches benefit from the number of valid problem generations for larger $K$, but their growth differs. The Iterative method exhibits a steeper rise and yields a consistently higher number of successful isomorphic problems across all request size while the growth of the Batch method tapers beyond $K=10$. Analyzing these outcomes suggests that, as K increases, the language model’s individual responses of the Batch method become longer and at times it refrains from generating the full set of requested problems, supplying fewer items than specified and omitting the remainder. In contrast, the right plot presents the complementary perspective of efficiency. Because a well-designed request prompt for isomorphic problem generation is typically much longer than the answer tokens (the generated problems themselves), the Iterative method requiring a fresh prompt for every single problem inevitably incurs a higher token cost as K increases. Our subsequent analysis revealed an additional factor that accentuates this gap; all Batch-generated problems originating from the same source either jointly passed or jointly failed the benchmark verification, indicating that the Batch method effectively constructs an internal generation engine first and then samples individual problems from this latent structure.

The left plot shows that while the Iterative approach consistently yields a higher number of successful problem generations as the requested number of problems increases. This advantage comes at the cost of growing inefficiency, illustrated in the right plot. Specifically, the Tokens Per Success (TPS) of the Iterative approach not only remains stagnant but worsens as the request size grows—showing a $9\%$ increase from $K=1$ to $K=10$, and a $22\%$ increase from $K=10$ to $K=30$.. This inefficiency is attributable to the growing likelihood of generating duplicate problems, which not only wastes tokens but also fails to meet the target number of distinct outputs.

The Batch approach, in contrast, exhibits remarkable efficiency gains as request size increases: TPS improves by $81\%$ when moving from $K=1$ to $K=10$, and by an additional $29\%$ from $K=10$ to $K=30$. This efficiency stems from the fact that a single prompt can generate multiple problems without proportionally increasing token consumption. Notably, the Batch approach also exhibits perfect internal consistency: for each original problem, all generated variants either pass or fail the benchmark verification as a group. This suggests that the Batch prompt effectively induces the LLM to construct an implicit “generation engine”—a latent representation of the underlying mathematical rules—before sampling concrete problem instances. This emergent behavior aligns with the goal of isomorphic problem generation, where a shared structure must be preserved across all outputs.

In summary, while the Iterative approach offers higher success counts in small-scale settings, its efficiency deteriorates and consistency falters as scale increases. The Batch approach, on the other hand, not only scales more efficiently but also reliably enforces structural uniformity, which is crucial for the industrial deployment of automated problem generation.

\clearpage
\section{Code-guided Prompting for CBIT}
\label{sec:appendix_cg}
\begin{figure}[h!]
  \includegraphics[width=\textwidth]{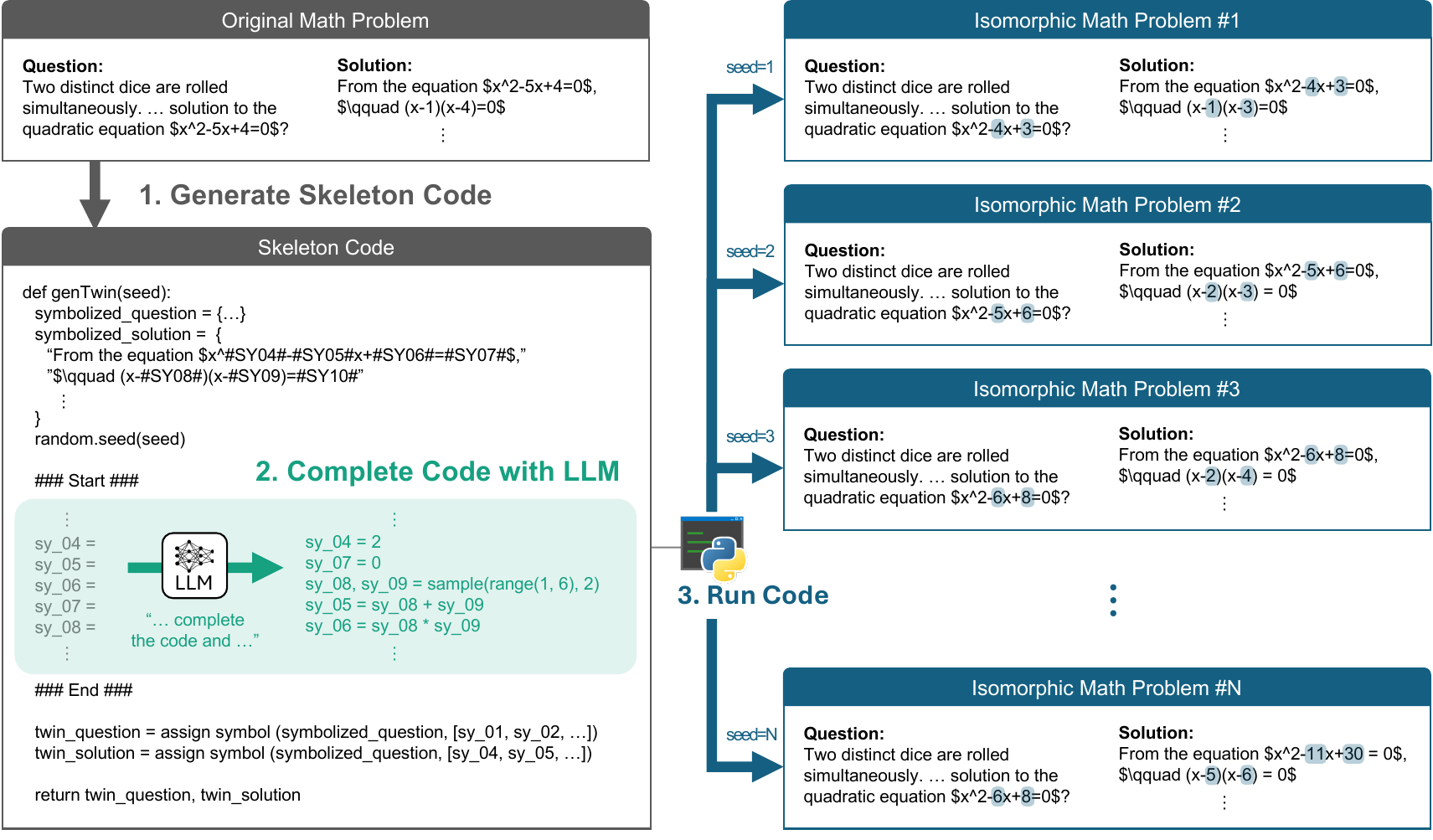}
  % \caption{Code-guided Prompting for CBIT}
  \caption{Illustration of CBIT’s meta-generation process. 
Given an original math problem, a skeleton code is first generated to symbolically represent its core relations. 
The LLM then completes the code by instantiating numerical values under the specified constraints, after which the executable program deterministically produces multiple isomorphic math problems from different random seeds. 
This design shifts the LLM’s role from directly authoring problems to constructing a generator, enabling scalable, consistent, and verifiable problem creation.}
  \label{fig:code_guided}
\end{figure}

\begin{figure}[h!]
\begin{center}
  \includegraphics[width=10cm]{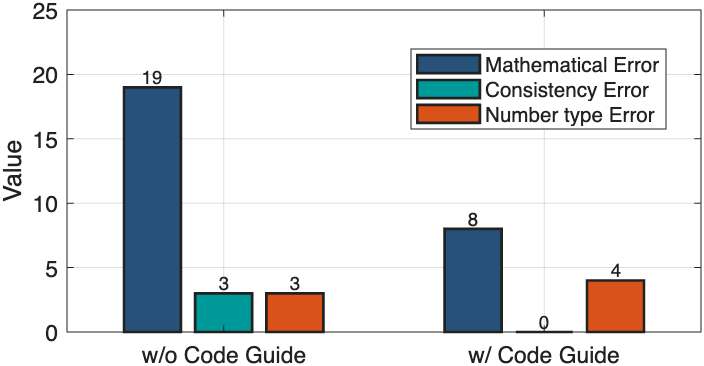}
  \caption {Impact of Skeleton Code on error reduction in CBIT. 
Providing Skeleton Code eliminates consistency errors entirely and substantially reduces both mathematical and number type errors by guiding the model’s symbol substitutions and structural alignment.}
  \label{fig:code_guide}
\end{center}
\end{figure}
\noindent

\end{document}